\documentclass{ecai} 
\usepackage{latexsym}
\usepackage{amssymb}
\usepackage{amsmath}
\usepackage{amsthm}
\usepackage{booktabs}
\usepackage{enumitem}
\usepackage{graphicx}
\usepackage{color}

\usepackage{multirow}
\usepackage[
singlelinecheck=false % <-- important
]{caption}

\newcommand{\BibTeX}{B\kern-.05em{\sc i\kern-.025em b}\kern-.08em\TeX}

\usepackage[disable]{todonotes}
\begin{document}

%%%%%%%%%%%%%%%%%%%%%%%%%%%%%%%%%%%%%%%%%%%%%%%%%%%%%%%%%%%%%%%%%%%%%%%%

\begin{frontmatter}

%%% Use this command to specify your submission number.
%%% In doubleblind mode, it will be printed on the first page.

\paperid{123} 

%%% Use this command to specify the title of your paper.

%\title{EvaluativeAI: A Hypothesis-Driven Decision Support Tool}

\title{Visual Evaluative AI: A Hypothesis-Driven Tool with Concept-Based Explanations and Weight of Evidence}

%%% Use this combinations of commands to specify all authors of your 
%%% paper. Use \fnms{} and \snm{} to indicate everyone's first names 
%%% and surname. This will help the publisher with indexing the 
%%% proceedings. Please use a reasonable approximation in case your 
%%% name does not neatly split into "first names" and "surname".
%%% Specifying your ORCID digital identifier is optional. 
%%% Use the \thanks{} command to indicate one or more corresponding 
%%% authors and their email address(es). If so desired, you can specify
%%% author contributions using the \footnote{} command.

\author[A]{\fnms{Thao}~\snm{Le}\thanks{Corresponding Author. Email: thaol4@student.unimelb.edu.au}}
\author[B]{\fnms{Tim}~\snm{Miller}}
\author[A]{\fnms{Ruihan}~\snm{Zhang}}
\author[A]{\fnms{Liz}~\snm{Sonenberg}}
\author[C]{\fnms{Ronal}~\snm{Singh}}

\address[A]{School of Computing and Information Systems, The University of Melbourne, Australia}
\address[B]{School of Electrical Engineering and Computer Science, The University of Queensland, Australia}
\address[C]{CSIRO/Data61, Australia}

%%% Use this environment to include an abstract of your paper.

\begin{abstract}
This paper presents \textbf{Visual Evaluative AI}, a decision aid that provides positive and negative evidence from image data for a given hypothesis. This tool finds high-level human concepts in an image and generates the Weight of Evidence (WoE) for each hypothesis in the decision-making process. We apply and evaluate this tool in the skin cancer domain by building a web-based application that allows users to upload a dermatoscopic image, select a hypothesis and analyse their decisions by evaluating the provided evidence. Further, we demonstrate the effectiveness of \textit{Visual Evaluative AI} on different concept-based explanation approaches.
\end{abstract}

\end{frontmatter}

%%%%%%%%%%%%%%%%%%%%%%%%%%%%%%%%%%%%%%%%%%%%%%%%%%%%%%%%%%%%%%%%%%%%%%%%

\section{Introduction}

\begin{figure}[t]
    \centering
    \caption{Comparison between the recommendation-driven paradigm and the hypothesis-driven paradigm}
    \includegraphics[width=\linewidth]{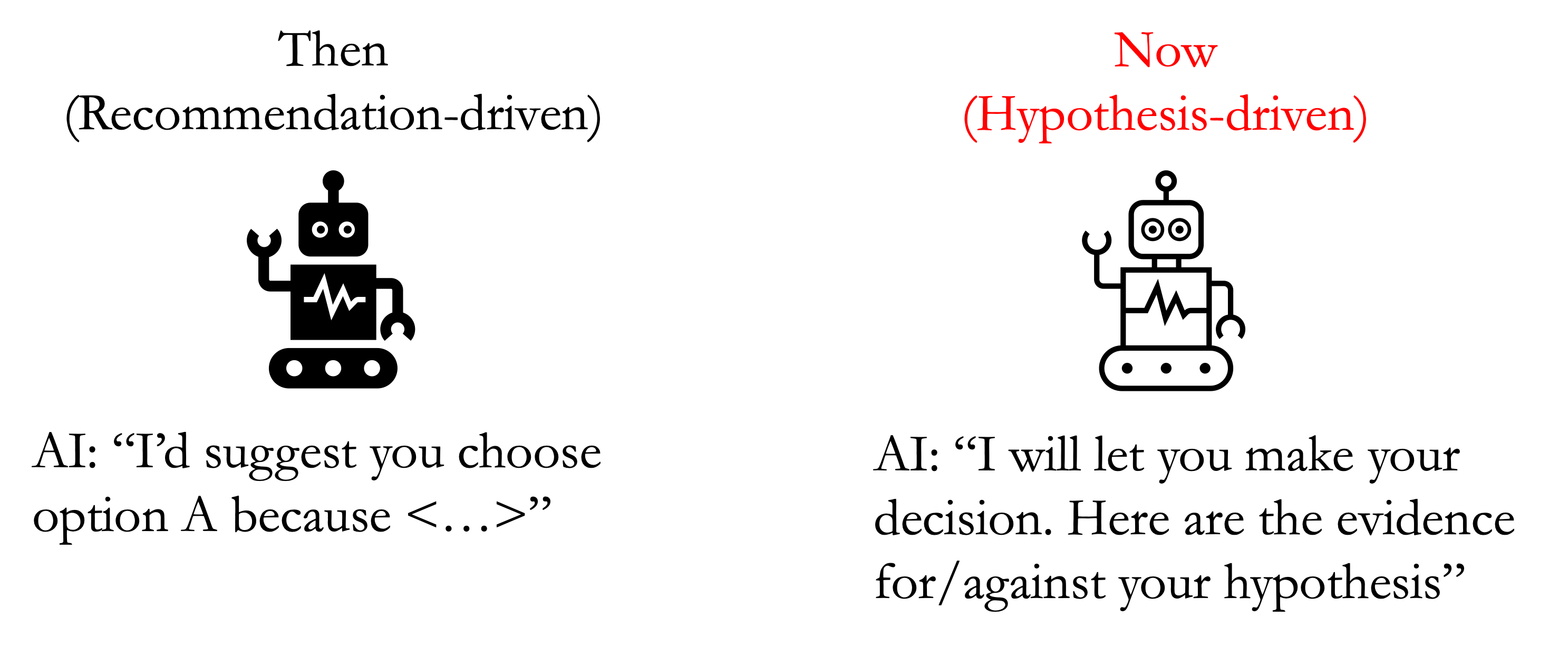}
    \label{fig:comparison-now-then}
\end{figure}

A common decision support paradigm called \textit{recommendation-driven} provides either or both the AI recommendation and the explanation for the given recommendation~\cite{Tschandl20,Bansal-complementary21,WangEffects22}. However, this paradigm is yet to be effective because it limits the control of human decision-makers, which can cause \textit{algorithm aversion}~\cite{dietvorst2015} where people do not trust the AI; or worse, \textit{over-reliance} on the AI system~\cite{vered2023}. \citet{miller2023explainable}  proposes a paradigm shift called \textbf{hypothesis-driven} using a conceptual framework \textbf{evaluative AI}. Figure~\ref{fig:comparison-now-then} describes the difference between the traditional recommendation-driven paradigm and the new hypothesis-driven paradigm. Rather than telling the decision-makers what to do, hypothesis-driven aims to reduce the reliance~\cite{le2024new}, promote uncertainty awareness~\cite{le2023dcexplaining,le2024new} and give decision-makers more control of the decision-making process by incorporating their hypotheses.

In this paper, we build an \textit{EvaluativeAI} tool by combining concept-based explanations for image data and the Weight of Evidence (WoE) model. This tool offers hypothesis-driven decision-making by generating evidence for possible hypotheses of an image. We also provide public access to \textit{EvaluativeAI} as a Python package so other researchers can use our tool. Moreover, we demonstrate a web-based application using this tool on a skin cancer dataset where users can see positive/negative evidence for different skin cancer diagnoses. The performance of the proposed models is evaluated on this skin cancer dataset.

\section{Methodology}

% \begin{figure}[ht]
%     \centering
%     \caption{Unlabelled concept (ICE+WoE)}
%     \includegraphics[width=\linewidth]{img/unlabelled-concept.png}
%     \label{fig:unlabelled-concept}
% \end{figure}

\begin{figure}[ht]
    \centering
    \caption{Labelled concept (PCBM+WoE)}
\includegraphics[width=\linewidth]{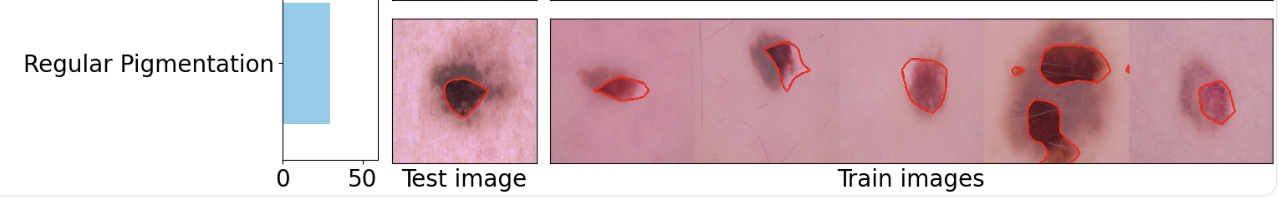}
    \label{fig:labelled-concept}
\end{figure}

\paragraph{Concept-based explanations} Concept-based models provide explanations using human-defined concepts that are related to parts of images~\cite{Kim18,Ghorbani2019concept}. The explanation is visualised as a segmentation of the image that represents a specific concept. The concept-based model can be classified into two categories: (1) supervised concept learning (concepts are labelled on each image in the training dataset) and (2) unsupervised concept learning (not having concept labels in the training dataset). Supervised concept learning requires labelled concepts in the training set, or the concepts can be transferred using another labelled dataset~\cite{yuksekgonul2023posthoc}. Unsupervised learning concept methods do not require the concepts to be labelled during the training process. This method is helpful when labelling concepts can be laborious, require expertise, or are not always available. Moreover, unsupervised learning can give users more agency as they can find a new concept that has not been labelled, but is still used by a machine learning model.

\paragraph{Weight of Evidence (WoE)} To measure a quantitative response of how much each concept (or feature) contributes in favour of, or against a particular hypothesis, we apply the Weight of Evidence (WoE) model~\cite{Melis-from-21}, building on~\citet{Good1985weight}. Through Bayes rule, WoE is expressed as the log-odd ratios of the evidence likelihood. In our application, the evidence will be referred to a concept (or feature) found in the image. Each concept will have a positive/negative quantitative value that shows how much it contributes to the given hypothesis.

We build our evidence generation model by combining a concept-based explanation model (i.e., Invertible Concept-based Explanation (ICE)~\cite{zhang2021improving}, Post-hoc Concept Bottleneck Model (PCBM)~\cite{yuksekgonul2023posthoc}) and the Weight of Evidence (WoE) model~\cite{Melis-from-21}. In particular, we replace the classifier layer of ICE and PCBM with the WoE model. Combining them together, we propose two models to generate the evidence-based explanations called \textit{ICE+WoE} and \textit{PCBM+WoE}. For example, Figure~\ref{fig:demo} and~\ref{fig:labelled-concept} demonstrate how the evidence is shown in the app. When we apply \textit{ICE+WoE}, the concept (or feature) does not have a label name, and is represented as a feature index (e.g., Feature 1 to 8). Furthermore, we need to choose the number of concepts with \textit{ICE}. Alternatively, \textit{PCBM+WoE} can provide a concept name (e.g., Regular Pigmentation) for each concept and the number of concepts is fixed based on the concept bank.

\section{Demonstration: A Case Study on Skin Cancer}

\begin{figure*}[!th]
    \centering
    \caption{Screenshot of \textit{Evaluative Skin Cancer} app when using unsupervised concept learning (ICE+WoE)}
    \includegraphics[width=\textwidth]{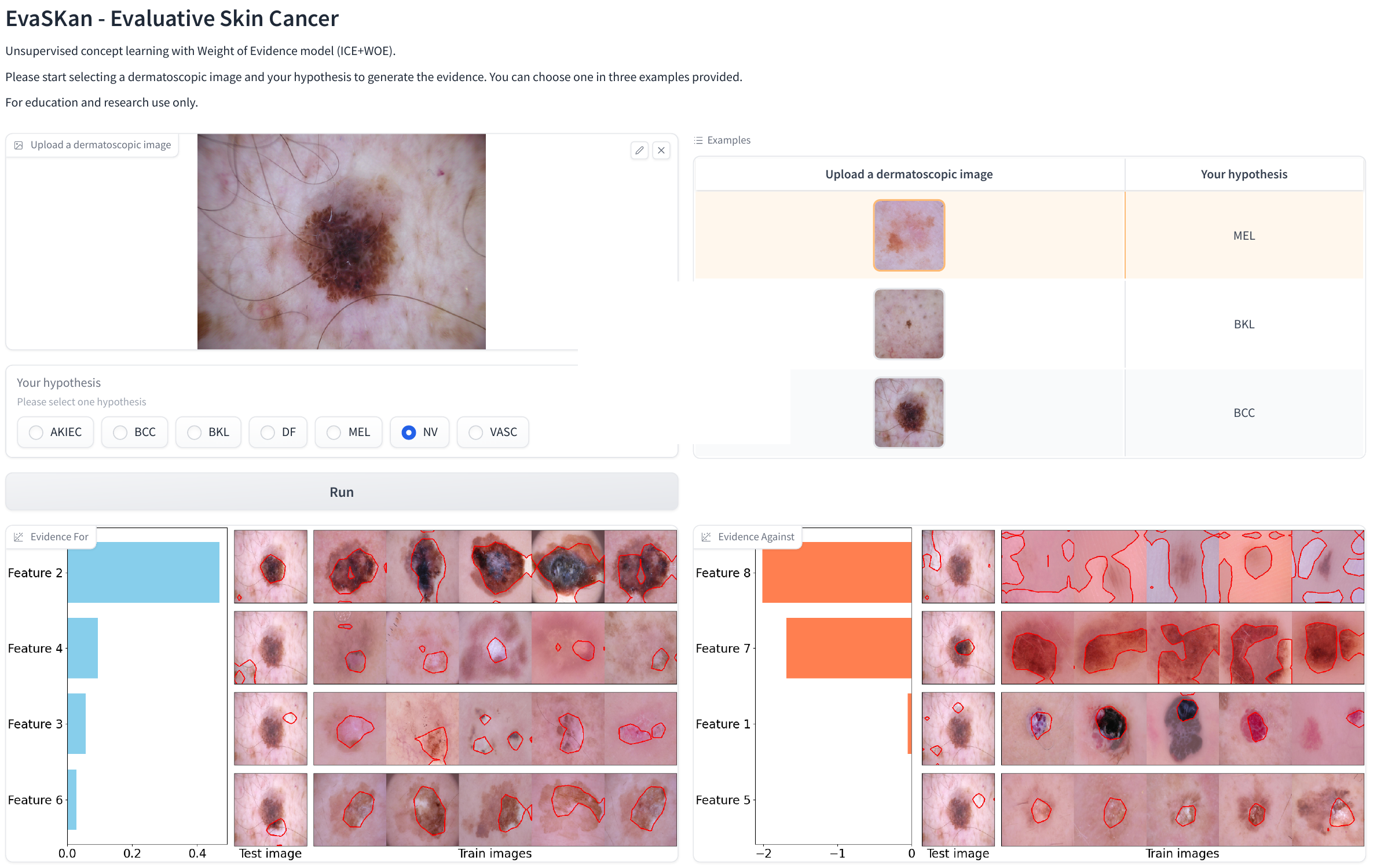}
    \label{fig:demo}
\end{figure*}

Applying AI in supporting skin cancer detection has become more prevalent and potentially improved decision-making accuracy. To demonstrate the effectiveness of \textit{Evaluative AI}, we apply this tool to the skin cancer diagnosis domain. We build a web-based application called \textit{Evaluative Skin Cancer (\textbf{EvaSkan})}, a solution for evaluating skin cancer using the hypothesis-driven paradigm. Users can select a hypothesis and the application will generate positive/negative evidence for that particular hypothesis. Using the evidence provided, the decision-maker can integrate their domain knowledge and make the final decision. Our application offers the foundation of an evaluative AI decision support tool (DST) in skin cancer diagnosis by focusing on human decision-makers, which is critical in the medical domain.

Figure~\ref{fig:demo} shows the user interface of the web-based application \textit{\textbf{EvaSkan}}. There are four main components in this app: (1) Upload a dermatoscopic image, (2) Three example test images, (3) Candidate hypotheses and (4) Evidence For/Against the selected hypothesis. First, the user can either select a dermatoscopic image of their choice or choose one test image in the three examples provided. Then, they select a hypothesis among the seven potential hypotheses/diagnoses~\footnote{https://challenge.isic-archive.com/landing/2018/47/}~\cite{Tschandl18}: AKIEC, BCC, BKL, DF, MEL, NV and VASC. When the user clicks \textit{Run}, the evidence for and against the selected hypothesis will be generated. Specifically, an image feature (concept) is described by an annotation in the selected test image and five other annotated examples in the training set that describe the same feature. For each feature, the app will show a quantitative measure of how much each feature provides in favour of, or against the selected hypothesis. By considering all possible hypotheses/diagnoses and the positive/negative evidence of the corresponding hypothesis, it is up to the user to make the final diagnosis, and they can choose to use the evidence from the DST. In our demonstration, users can try both the unlabelled concept approach (i.e., ICE+WoE) and the labelled concept approach (i.e, PCBM+WoE). Figure~\ref{fig:demo} shows an example when using unsupervised concept learning (ICE+WoE).  

\section{Experiments}

\subsection{Dataset and Model Implementation} We use the \textbf{HAM10000 dataset}~\cite{Tschandl18} to train all models (original CNN backbones, ICE, ICE+WoE, PCBM and PCBM+WoE). The data has a total of 10015 dermatoscopic images and seven output classes: Actinic keratoses (AKIEC), basal cell carcinoma (BCC), benign keratosis (BKL), dermatofibroma (DF), melanoma (MEL), melanocytic nevi (NV) and vascular lesion (VASC). We balanced the dataset by applying Weighted Random Sampler~\footnote{\url{https://pytorch.org/docs/stable/_modules/torch/utils/data/sampler.html}} and data augmentation. Finally, each class has 1000 samples that were used for the training process, making a total of 7000 samples for seven classes. The test set is selected as a fraction of the original dataset (without augmentation). As in the original HAM10000, class DF has the lowest number of samples (i.e., 75 samples). Therefore, we choose 20 samples in each class for the test set, which represents 26\% of class DF. We then have a total of 140 samples for the test set to evaluate the model performance.

Since images in the HAM10000 dataset do not have the concept labels,  to get the concept labels for the PCBM model, we train Concept Activation Vectors (CAVs)~\cite{Kim18} on the \textbf{7-point checklist dataset}~\cite{kawahara2018seven}. Followed the previous work~\cite{yuksekgonul2023posthoc,Yan23towards}, we have 12 concepts: \textit{Atypical Pigment Network}, \textit{Typical Pigment Network}, \textit{Blue Whitish Veil}, \textit{Irregular Vascular Structures}, \textit{Regular Vascular Structures}, \textit{Irregular Pigmentation}, \textit{Regular Pigmentation}, \textit{Irregular Streaks}, \textit{Regular Streaks}, \textit{Regression Structures}, \textit{Irregular Dots and Globules} and \textit{Regular Dots and Globules}. The PCBM model then used the trained CAVs based on these 12 concepts and applied that to extract the concept. For each concept, we have 50 positive samples (contain the concept) and 50 negative samples (do not contain the concept). The learning rate was set to $0.01$ and ridge regression was used at the classifier layer of PCBM. 

\subsection{Results}

\begin{figure}
    \caption{F1-score of ICE, ICE+WoE and the original ResneXt50 over different number of concepts. On the right, we  compare ICE using different reducers NMF and PCA.}
    \noindent\begin{minipage}[b]{0.33\linewidth}
		\centering
        \includegraphics[width=\linewidth]{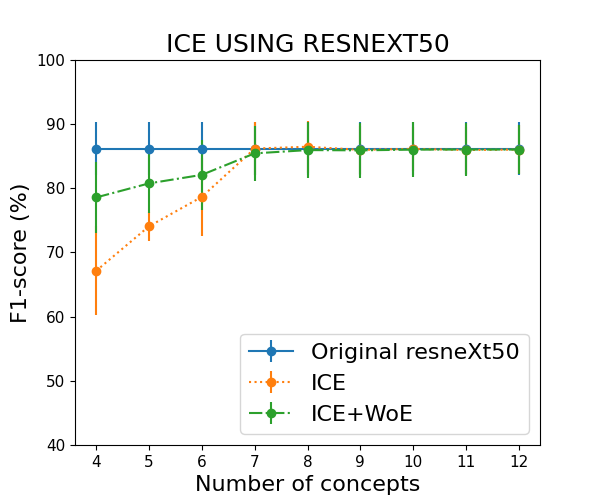}
    \end{minipage}
    \noindent\begin{minipage}[b]{0.33\linewidth}
    	\centering
        \includegraphics[width=\linewidth]{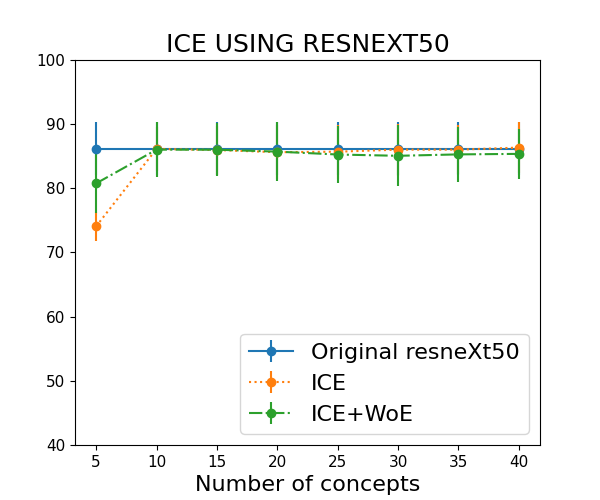}
    \end{minipage}
    \noindent\begin{minipage}[b]{0.326\linewidth}
    	\centering
        \includegraphics[width=\linewidth]{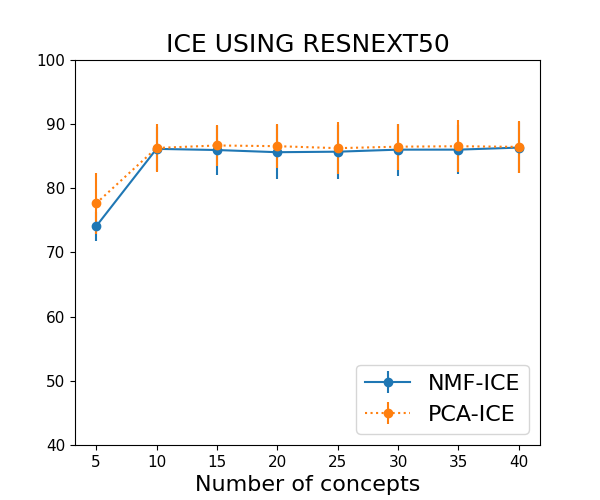}
    \end{minipage}
    \label{fig:rx50-by-concept}
\end{figure}

\begin{table}
    \centering
    \setlength{\tabcolsep}{1.8pt}
    \caption{Performance for the original CNN model, ICE, ICE+WoE, PCBM and PCBM+WoE. The ICE model uses an NMF (non-negative matrix factorization) reducer. ICE(7) represents the ICE model with 7 different concepts. PCBM(12) is the PCBM model with 12 labelled concepts. \textit{mean} $\pm$ \textit{standard deviation} of the performance are reported over 20 random seeds. Winners are indicated in bold.}
    \label{tab:evaskan-computational}
    \begin{tabular}{lllll}
        \toprule
        CNN & Model & Precision $\uparrow$ & Recall $\uparrow$ & F1-Score $\uparrow$ \\
        Backbone & & & &\\
        \midrule
        \multirow{5}{*}{Resnet50}
        & Backbone & $83.08 \pm 5.98$ & $85.33 \pm 6.20$ & $\mathbf{84.04 \pm 5.01}$ \\
        & ICE(7) & $73.34 \pm 8.69$ & $87.50 \pm 10.04$ & $78.99 \pm 4.91$ \\
        & ICE(7)+WoE & $80.13 \pm 5.44$ & $82.00 \pm 6.81$ & $80.85 \pm 4.55$ \\
        & PCBM(12) & $73.93 \pm 8.94$ & $82.08 \pm 12.67$ & $76.58 \pm 6.31$ \\
        & PCBM(12)+WoE & $80.73 \pm 5.21$ & $84.25 \pm 3.35$ & $82.32 \pm 2.98$ \\
        \midrule
        \multirow{5}{*}{ResneXt50}
        & Backbone & $85.46 \pm 4.63$ & $87.25 \pm 6.31$ & $86.20 \pm 4.18$ \\
        & ICE(7) & $84.23 \pm 5.49$ & $88.58 \pm 5.41$ & $\mathbf{86.20 \pm 4.11}$ \\
        & ICE(7)+WoE & $84.73 \pm 5.00$ & $86.33 \pm 4.76$ & $85.45 \pm 4.25$ \\
        & PCBM(12) & $78.93 \pm 8.28$ & $83.17 \pm 14.43$ & $79.83 \pm 8.28$ \\
        & PCBM(12)+WoE & $84.48 \pm 4.86$ & $85.50 \pm 3.98$ & $84.92 \pm 3.64$ \\
        \midrule
        \multirow{5}{*}{Resnet152}
        & Backbone & $84.49 \pm 6.48$ & $86.08 \pm 5.70$ & $\mathbf{84.96 \pm 3.09}$ \\
        & ICE(7) & $78.30 \pm 8.11$ & $87.42 \pm 7.48$ & $82.10 \pm 4.37$ \\
        & ICE(7)+WoE & $81.21 \pm 4.90$ & $85.08 \pm 5.14$ & $83.01 \pm 4.13$ \\
        & PCBM(12) & $76.49 \pm 7.75$ & $87.08 \pm 5.15$ & $81.09 \pm 4.21$ \\
        & PCBM(12)+WoE & $82.97 \pm 5.37$ & $84.83 \pm 4.04$ & $83.73 \pm 2.99$ \\
        \bottomrule
    \end{tabular}
\end{table}

\paragraph{ICE+WoE and PCBM+WoE achieve comparable performance to the original CNN models} Table~\ref{tab:evaskan-computational} reports the performance of ICE(7), ICE(7)+WoE, PCBM(12) and PCBM(12)+WoE using three different CNN backbone models (Resnet50, Resnet152~\cite{Kaiming16} and ResneXt50~\cite{Saining17}). We select 12 concepts for PCBM based on previous work~\cite{yuksekgonul2023posthoc,Yan23towards}. For ICE, we run experiments with the number of concepts ranging from 5 to 40. As shown in Figure~\ref{fig:rx50-by-concept}, performance peaks at 7 concepts. Therefore, the final comparison in this table is made between ICE(7) and PCBM(12).
    
The results show that ICE(7)+WoE and PCBM(12)+WoE achieve comparable performance to the original CNN models. Particularly, with ResneXt50, the F1-score of ICE(7)+WoE and PCBM(12)+WoE are $85.45 \pm 4.25$ and $84.92 \pm 3.64$, respectively, while the original ResneXt50 has an F1-score of $86.20 \pm 4.18$. Therefore, ICE(7)+WoE (using 7 features) and PCBM(12)+WoE (using 12 features) show comparable performance compared to the original ResneXt50 with 2048 features. Moreover, similar to the findings in~\cite{zhang2021improving}, when we compare the performance using different reducers as in Figure~\ref{fig:rx50-by-concept}, NMF and PCA (principal component analysis), PCA provided the best performance but could be less interpretable compared to NMF.
 
\paragraph{Having more concepts did not lead to better accuracy} Figure~\ref{fig:rx50-by-concept} shows the performance of the original ResneXt50, ICE and ICE+WoE over different numbers of concepts from 5 concepts to 40 concepts. Two figures from the left show the performance of ICE using the NMF reducer. When there are 5 concepts, ICE(5)+WoE ($80.43 \pm 4.60$) has a significantly higher F1-score than ICE(5) ($74.08 \pm 2.26$) ($p=2.41 \times 10^-6 < 0.001$, $d=1.753$). Since we have 2048 features at the classifier layer of ResneXt50, ResneXt50 outperforms ICE(5)+WoE and ICE(5) significantly ($p<0.001$). But the performance of both ICE+WoE and ICE match the performance of the original ResneXt50 when we have at least 7 concepts. Particularly, with \textit{as few as 7 concepts}, ICE and ICE+WoE achieve similar performance to the original ResneXt50 using 2048 features. The performance of ICE and ICE+WoE also stopped improving at 7 concepts with a backbone of ResneXt50. The reason is that when we apply a reducer in ICE (e.g. NMF), some important concepts are detected at first. Then after we increase the number of concepts, some noisy concepts are detected, which could lead to a slight drop in the performance. Eventually, all important concepts are found and match the performance of the original CNN model.

In summary, the results show that with a few number of concepts (i.e., 7 concepts), we can achieve comparable performance compared to the original CNN models. Therefore, this indicates the accuracy of the evidence being generated, which is potentially useful to the decision-makers. Importantly, despite the concept-based models (ICE(7), ICE(7)+WoE, PCBM(12) and PCBM(12)+WoE) being slightly less accurate than the CNN backbones, it would also be much easier for users to interpret and evaluate the evidence by not showing too many concepts.

\section{Conclusion}
In this paper, we introduce \textbf{Visual Evaluative AI}~\footnote{\url{https://github.com/thaole25/EvaluativeAI}}, a tool for hypothesis-driven decision support. This tool can highlight the high-level concepts in an image and provide positive/negative evidence for all possible hypotheses. Our tool is further applied and evaluated in the skin cancer domain with a web-based application called \textit{\textbf{EvaSKan}} that offers skin cancer diagnosis support. In future work, a more comprehensive evaluation will be undertaken by addressing the domain expert opinions on this application.

%%%%%%%%%%%%%%%%%%%%%%%%%%%%%%%%%%%%%%%%%%%%%%%%%%%%%%%%%%%%%%%%%%%%%%%%

%%% Use this environment to include acknowledgements (optional).
%%% This will be omitted in doubleblind mode.

\begin{ack}
This research was supported by \textit{(a)} the University of Melbourne Research Scholarship (MRS), \textit{(b)} Australian Research Council Discovery Grant DP190103414 and undertaken using the \textit{(c)} LIEF HPC-GPGPU Facility hosted at the University of Melbourne. This Facility was established with the assistance of LIEF Grant LE170100200.
\end{ack}

%%%%%%%%%%%%%%%%%%%%%%%%%%%%%%%%%%%%%%%%%%%%%%%%%%%%%%%%%%%%%%%%%%%%%%%%

%%% Use this command to include your bibliography file.

\bibliography{references}

\end{document}